\documentclass{article}
\usepackage{spconf,amsmath,graphicx}
\usepackage{multirow}
\usepackage{booktabs}
\usepackage{color}
\usepackage{marvosym}
\usepackage{pifont}

\usepackage[normalem]{ulem}
\useunder{\uline}{\ul}{}

\title{Dynamic data sampler for cross-language transfer learning in large language models}
\name{Yudong Li$^{ 1}$ \quad Yuhao Feng$^{2}$ \quad Wen Zhou$^{3}$ \quad Zhe Zhao$^{2}$ \quad Linlin Shen$^{1*}$  \quad Cheng Hou$^{2}$ \quad Xianxu Hou$^{4}$}
  
\address{$^{1}$ School of Computer Science and Software Engineering, Shenzhen University \\ $^{2}$Tencent AI Lab \\ $^{3}$ LIESMARS, Wuhan University  \\ $^{4}$School of AI and Advanced Computing, Xi’an Jiaotong-Liverpool \thanks{*Corresponding author: llshen@szu.edu.cn}}

\begin{document}
%\ninept
%
\maketitle
\begin{abstract}
Large Language Models (LLMs) have gained significant attention in the field of natural language processing (NLP) due to their wide range of applications. However, training LLMs for languages other than English poses significant challenges, due to the difficulty in acquiring large-scale corpus and the requisite computing resources. In this paper, we propose ChatFlow, a cross-language transfer-based LLM, to address these challenges and train large Chinese language models in a cost-effective manner. We employ a mix of Chinese, English, and parallel corpus to continuously train the LLaMA2 model, aiming to align cross-language representations and facilitate the knowledge transfer specifically to the Chinese language model. In addition, we use a dynamic data sampler to progressively transition the model from unsupervised pre-training to supervised fine-tuning. Experimental results demonstrate that our approach accelerates model convergence and achieves superior performance. We evaluate ChatFlow on popular Chinese and English benchmarks, the results indicate that it outperforms other Chinese models post-trained on LLaMA-2-7B.
\end{abstract}
\section{Introduction}
\label{sec:intro}

Large Language Models (LLMs) have demonstrated significant application potential and become a research hotspot. At present, there are many publicly available models that provide robust baselines, which promote the research of LLMs within the NLP community. 
Most of these models benefit from the naturally high-quality English corpus available on the Internet. For instance, models such as  LLaMA1\cite{touvron2023llama}, Falcon\cite{refinedweb} and LLaMA2\cite{touvron2023llama2}, are pre-trained on large-scale corpus (exceeding 1000B tokens) and have achieved state-of-the-art performance.

However, training LLMs in languages other than English presents great challenges. For example, by August 2023, English text accounted for 59.3\% of the content on the Internet, while Chinese made up only 1.4\% \footnote{https://w3techs.com/technologies/overview/content\_language}. This significant disparity makes it particularly difficult to train LLMs of comparable scale and quality in the Chinese language. The existing Chinese public models like ChatGLM\cite{zeng2022glm}, Baichuan \footnote{https://github.com/baichuan-inc/Baichuan-13B} and Qwen\footnote{https://github.com/QwenLM/Qwen-7B} are typically pre-trained on their private data without disclosing training details, which poses challenges for reproducibility.
In addition, training such models requires significant engineering capabilities across multiple stages including pre-training, supervised fine-tuning (SFT), reinforcement learning from human feedback (RLHF), and requires substantial computing resources and manpower. This overhead makes great challenges to train LLMs from scratch for most researchers.

To alleviate these issues and explore a cost-effective method for constructing large Chinese language models, we propose ChatFlow, a cross-language transfer-based LLM. 
We conduct continuous training on an English-specific language model using a mix of Chinese, English and parallel corpus to align the cross-language representation, which allows for transferring the English model's inherent knowledge to Chinese. 
Moreover, inspired by curriculum learning\cite{bengio2009curriculum}, we propose a dynamic data sampler that progressively transition the model from unsupervised pre-training to supervised fine-tuning using training data with dynamic distribution.
In contrast to existing methods that separate pre-training and fine-tuning stages, our approach provides a smoother transition for model training, avoiding abrupt changes in data distribution between different stages. %Experimental results show that our method accelerates model convergence and achieves superior performance.

We train our proposed ChatFlow on approximately 50GB data based on the LLaMA2-7B foundation model. 
We find that our model can readily learn Chinese knowledge while retaining its original English capability.
We evaluate ChatFlow on popular Chinese and English benchmarks. In automatic evaluations including MMLU\cite{hendrycks2020measuring}, C-Eval\cite{huang2023ceval}, CMMUL\cite{li2023cmmlu} and GAOKAO\cite{zhang2023evaluating}, compared with other Chinese models post-trained on LLaMA-2-7B, our model achieves superior results. In the human-based SuperCLUE\cite{xu2023superclue} benchmark, our model ranks 5-th among 7B-scale models. Notably, unlike other Chinese-native models, ChatFlow is trained from an English foundation model and requires less Chinese data. Furthermore, we only used publicly available data for training and have released both the code and weights\footnote{https://github.com/CVI-SZU/Linly} for reproducibility. 
%Although ChatFlow's overall performance does not surpass commercial closed-source LLMs, it provides a reproducible baseline and is a valuable training LLMs cross different languages.

Our contributions are summarized as follows:

\begin{itemize}
\item We propose ChatFlow, a novel transfer-based large language model that enables cost-effective training of cross-language LLMs.

\item We introduce a dynamic data sampler, which removes the explicit gap between pre-training and SFT.% which, facilitating smoother distribution spanning between the pre-training and instruction-tuning phases.
\item We train our ChatFlow based on LLaMA-7B to boost its Chinese performance, and experimental results demonstrate its superiority over other methods.
\end{itemize}

%\section{Related Work}

%\subsection{Alpaca}

%\subsection{Chinese Models}

\section{Approach}

\subsection{Transfer Learning with Dynamic Data Sampler}

% 通常，训练Chat模型需要在LLM上进行SFT。这种训练学习如何回复人类的提问，被视为一种迁移学习。然而，这种训练无法大规模进行，因为分布的变化容易导致LLM遗忘原本的知识，

%Existing LLM training typically involves separate stages of unsupervised pre-training and supervised instruction-tuning. This process starts with pre-training, and upon its completion, the saved weights are loaded to initiate instruction-tuning using instruction data.

The conventional approach of large language model training typically involves separate stages of unsupervised pre-training and supervised fine-tuning (SFT). 
This design is based on the fact that the model acquires general knowledge during pre-training, while SFT focuses on transferring this knowledge to downstream tasks by learning the format of user interactions.
In previous practices, the sudden shift in data distribution during the SFT stage could cause the model to confuse previously learned knowledge. To mitigate this issue, unsupervised data are mixed with instructions to balance the distribution\cite{refinedweb}.

However, in training ChatFlow, both cross-language transfer and downstream task transfer take place concurrently, which amplifies the challenge of maintaining stable transfer learning.
In this work, inspired by curriculum learning\cite{bengio2009curriculum}, we employ a dynamic data sampler to facilitate a smoother transition in model training. For ChatFlow training, this approach transitions the model from English pre-training to bilingual (English and Chinese) pre-training and instruction-tuning in a stepwise manner, thereby speeding up convergence and improving performance. 

% 在我们的方法中，面料了更大的挑战。

%In this paper, we propose a training approach grounded with curriculum learning\cite{bengio2009curriculum} that provides a smoother transition. For our training, it transitions the model from English pre-training to bilingual (English and Chinese) pre-training and instruction-tuning in a stepwise manner, thereby speeding up convergence and improving performance. 

Specifically, we utilize a sampler to construct training batches. For each task, the function $\gamma(t)$ computes the sampling rate for the $t$-th sample.
Initially, the sampler applies a higher proportion of English and parallel corpus, mirroring the distribution of the LLaMA's original pre-training. As the training progresses, the sampler linearly increases the proportion of Chinese and instruction data. This transition is completed within $T_{grow}=5 M$ samples, indicating a shift in training from English to Chinese, and from unsupervised tasks to instruction learning. For the remaining training samples, a fixed distribution is used for consistent learning and improvement throughout the later stages of training. The function $\gamma(t)$ is defined as follows:

$$\gamma(t)=
\begin{cases}
\alpha + \frac{\beta - \alpha }{T_{grow}} \cdot t & {t} \le T_{grow} \\
\beta  & t > T_{grow} 
\end{cases}$$

\noindent where $\alpha$ represents the initial weight, $\beta$ is the final weight. We empirically set these parameters based on previous curriculum learning methods' experience. The settings for each task are presented in Table~\ref{datasets}.

\begin{table}[]
\caption{The properties of the mixed datasets used in the transfer learning and their hyperparameters. Lang: Language; inst: instruction.}
\label{datasets}
\resizebox{1.00\columnwidth}{!}{
\begin{tabular}{@{}llllll@{}}
\toprule
Dataset & Disc Size & Lang. & Type & $\alpha$ & $\beta$ \\ \midrule
RefinedWeb & 10 GB & en & corpus & \textit{0.60} & \textit{0.15} \\ \addlinespace
CLUECorpus & 13 GB & \multirow{3}{*}{zh} & \multirow{3}{*}{corpus} & \multirow{3}{*}{\textit{0.05}} & \multirow{3}{*}{\textit{0.50}} \\
WuDao & 10 GB &  &  &  &  \\
CSL & 1.5 GB &  &  &  &  \\ \addlinespace
ParaCrawl v9 & 2.6 GB & \multirow{2}{*}{en-zh} & \multirow{2}{*}{parallel} & \multirow{2}{*}{\textit{0.25}} & \multirow{2}{*}{\textit{0}} \\
WikiMatri & 0.6 GB &  &  &  &  \\ \addlinespace
UltraChat & 5 GB & \multirow{2}{*}{en} & \multirow{2}{*}{inst.} & \multirow{2}{*}{\textit{0.05}} & \multirow{2}{*}{\textit{0.10}} \\
FLAN & 1.7 GB &  &  &  &  \\ \addlinespace
BELLE & 4.6 GB & \multirow{2}{*}{zh} & \multirow{2}{*}{inst.} & \multirow{2}{*}{\textit{0}} & \multirow{2}{*}{\textit{0.20}} \\
COIG & 4.5 GB &  &  &  &  \\ \addlinespace
GitHub & 2 GB & multi & code & \textit{0.05} & \textit{0.05} \\ \bottomrule
\end{tabular}}
\vspace{-0.25cm}
\end{table}

\subsection{Training Data}

Our training data is composed of several data sources, including unsupervised corpus, parallel corpus, and instruction data in Chinese and English language. We use about 50GB data containing 8B tokens. The data source details are as follows:

%\begin{figure}[t]
%\centerline{\includegraphics[width=8cm]{figures/data.png}}
%\caption{Distribution over training data.}
%\label{dataset}
%\vspace{-0.25cm}
%\end{figure}

\textbf{Parallel corpus.}
The Chinese-English parallel corpus bridges Chinese and English knowledge representation within the model. By leveraging the foundation model's built-in English ability, parallel corpus accelerates the model's learning of Chinese knowledge.
We use ParaCrawl v9 \cite{espla2019paracrawl} and WikiMatri \cite{schwenk2021wikimatrix} to align sentence- and word-level Chinese-English representations.

\textbf{Unsupervised corpus.}
Our unsupervised data includes both Chinese and English corpus, where Chinese data is used to provide Chinese world knowledge, and English data is used to balance the training distribution to avoid forgetting of existing knowledge. The Chinese corpus contains CLUECorpus\cite{xu2020cluecorpus2020}, CSL\cite{li2022csl}, and a subset of WuDao dataset\footnote{https://github.com/BAAI-WuDao/Data} filtered by URL domain. The English corpus contains a subset of RefinedWeb\cite{refinedweb}. The GitHub code is obtained from SlimPajama \footnote{https://huggingface.co/datasets/cerebras/SlimPajama-627B}.

\textbf{Instruction data.}
The model acquires the ability to interact with users and enhances its knowledge by learning from instructional data. 
We combine instruction datasets from different sources and languages, including self-instructed data such as BELLE\cite{BELLE} and UltraChat\cite{ding2023enhancing}; supervised data and prompts such as FLAN\cite{weifinetuned}, COIG\cite{zhang2023chinese}.

\subsection{Prompt Format and Objective}

We train the ChatFlow model using a consistent language model objective, where unsupervised data and instruction data are distinguished by a specific format.
For each training instance, we directly use the unsupervised data for training. The parallel corpus, on the other hand, is spliced by line breaks and then used as a training sample. As for the instruction data, we encapsulate it within a prompt template. The prompt template follows the Alpaca format\footnote{https://github.com/tatsu-lab/stanford\_alpaca} and is simplified. Multiple rounds of dialogue are separated by ``User'' and ``Bot'' symbols. The following is an example of a 2-round instruction:

\begin{quote}\em\small
User: \{question-1\} Bot: \{answer-1\} \#\#\# Instruction: \{question-2\} \#\#\# Response: \{answer-2\}

\end{quote}

We splice (or truncate) each instance into fixed-length sequences using the full-sentence strategy. Given a sequence of tokens, denoted as $\{t_1,...,t_n\}$ and a model with trainable parameters  $\theta$, we use the language modeling objective to maximize the following likelihood:

\begin{equation}
    \mathit{L(\theta)} = 
    \sum_{i} \textup{log}\mathit{P}\left ( t_i | t_1,...,t_{i-1}; \theta \right )
\end{equation}

\begin{table}[t]
\caption{A comparison of ChatFlow with LLaMA2-Chat and other Chinese models post-trained on LLaMA2.}
\label{your-label}
\resizebox{1.00\columnwidth}{!}{
\begin{tabular}{lcccc}
\toprule
Model & MMLU & C-Eval & CMMLU & GAOKAO \\ \midrule
%Meta-LLaMA2 & 40.4 & 29.2 & 30.8 & 27.1 \\
Meta-LLaMA2-Chat & 45.3 & 31.7 & 32.1 & 27.3 \\
FlagAlpha-LLaMA2 & 46.1 & 33.3 & \textit{32.8} & \textit{28.5} \\
LinkSoul-LLaMA2-sft & \textbf{47.4} & 34.5 & \textit{36.5} & \textit{30.2} \\
HFL-Alpaca2 & 43.0 & {\ul 39.9} & {\ul 39.4} & {\ul 36.2} \\
wzhu-LLaMA2-sft & 44.7 & 33.6 & 32.9 & 29.4 \\ \midrule
ChatFlow (ours) & {\ul 46.8} & \textbf{43.1} & \textbf{40.2} & \textbf{40.7} \\ \bottomrule
\label{cmp}
\end{tabular}}
\vspace{-0.5cm}
\end{table}

\section{Experiment}

\begin{figure}[t]
	\centering
	\begin{minipage}{1\linewidth}
		\centering
		\includegraphics[width=0.9\linewidth]{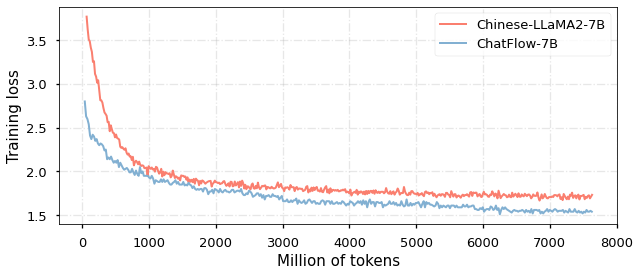}
		\caption{Training loss over the number of trained tokens in the ablation study.} %\medskip
            \vspace{0.5cm}
		\label{loss}%文中引用该图片代号
	\end{minipage}
	%\qquad
	%让图片换行，

	\begin{minipage}{1\linewidth}
		\centering
		\includegraphics[width=0.9\linewidth]{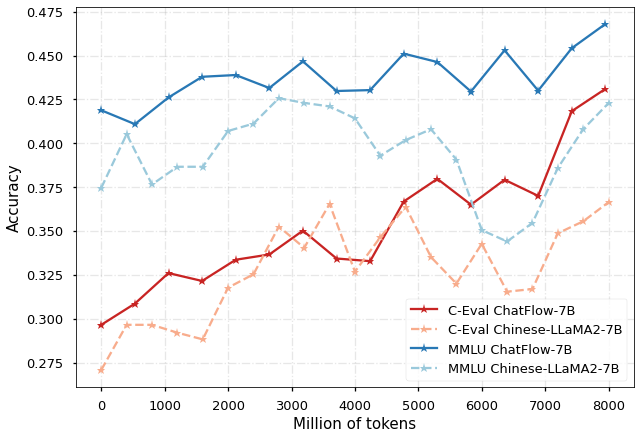}
		\caption{Evaluation metrics over trained tokens in the ablation study. The model's performance on the English evaluation MMLU is shown on the \textcolor{blue}{blue line}, while its performance on the Chinese evaluation C-Eval is shown on the \textcolor{red}{red line}.} %\medskip
		\label{exp1}%文中引用该图片代号
	\end{minipage}
\vspace{-0.25cm}
\end{figure}

\subsection{Training ChatFlow}

We initialize our model with LLaMA2-7B weights. Since LLaMA was originally designed for the English language, there are only 700 Chinese characters in its vocabulary. We extend the vocabulary with 8,701 Chinese characters and 62 symbols. We initialize the extended embedding and output matrices with the mean values of tokens corresponding to these words in the original vocabulary.

We train ChatFlow on 16*A100 GPUs for two weeks, using TencentPretrain framework\cite{zhao2022tencentpretrain} in bfloat16 format. We basically follow Meta's training hyperparameters. The sequence length is set to 2048, the batch size is 512, and gradient accumulation is used.

% 训练参数

\subsection{Main Results}

\begin{figure}[t]
\centerline{\includegraphics[width=8cm]{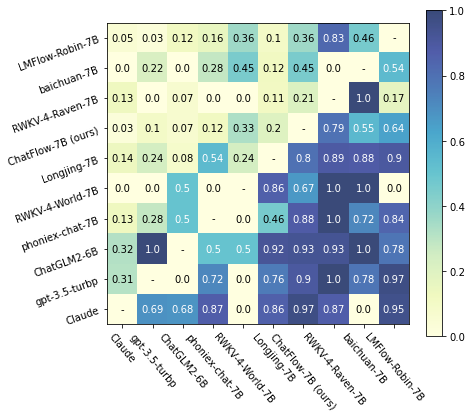}}
\caption{Win rate for all models in non-tie matches. ChatFlow ranks 5th among the 7B models.}
\label{clue}
%\vspace{-0.5cm}
\end{figure}

We evaluate our model on different benchmarks including MMLU\cite{hendrycks2020measuring} for English understanding, C-Eval\cite{huang2023ceval} for comprehensive Chinese understanding, CMMLU\cite{li2023cmmlu} for Chinese knowledge and reasoning and GAOKAO\cite{zhang2023evaluating} for Chinese college entrance examination questions. All evaluations are conducted under 3-shot setting. In Table~\ref{cmp}, we compare our model with other LLaMA2-based Chinese models. It can be seen that our model outperforms other models in Chinese and retains its English proficiency after post-training in Chinese. On the other hand, ChatFlow also reflects the efficiency of transfer learning. For example, compared with the previous SOTA model HFL-Alpaca2 which uses 120GB corpus for training, ChatFlow only uses less than half of its data.

% 在另一方面，ChatFlow也体现了迁移的准确性，
%We hypothesize that this can be attributed to curriculum learning, which helps mitigate disturbances caused by changes in data distribution. [TODO]

\subsection{Human Evaluation}

We also conduct a user-based study to evaluate ChatFlow, utilizing SuperCLUE \cite{xu2023superclue}, an anonymous competition platform designed for large Chinese models. Users are asked to chat with two randomly selected models, choosing the one they deem superior. The platform has collected 9.9k user votes, with model scores and rankings determined by the Elo scoring, which is a widely used rating system in chess and other competitive games.

We compare ChatFlow with state-of-the-art commercial models (including Claude and gpt-3.5-turbo), as well as other models with similar parameter levels.
The winning rate and detailed comparison are shown in Figure~\ref{clue} and Table~\ref{exp-2}. It can be seen that ChatFlow ranks 5th among the 7B models. Notably, ChatGLM2, RWKV-Word and Longjing are trained from scratch in both Chinese and English. Phoniex, on the other hand, is fine-tuned based on the BLOOMZ multilingual foundation model. Unique among these, ChatFlow is the only model based on an English model and acquiring Chinese language capabilities through transfer learning.
Despite these achievements, when compared with recent popular commercial models, our model still has a large gap, the winning rate is only about 10\%, which leaves room for future improvement.
%As it is shown in Figure~\ref{clue}, our ChatFlow ranks 12th in the overall ranking. Compared with state-of-the-art commercial models (including Claude and gpt-3.5-turbo), the winning rate is only about 10\%. Among models with comparable parameters (7B and below), our model ranks 5th.

\subsection{Ablation Study}

\begin{table}[t]
\caption{Comparing Chatflow to other models on human evaluation. An asterisk ``*'' indicates that the model was trained from scratch. Avail: Model availability, "API" denotes that the model can only be called through API, "Weights" denotes that the model weights are released, while the training data and code are not available.}
\label{exp-2}
\resizebox{1.00\columnwidth}{!}{
\begin{tabular}{lllll}
\toprule
Model & Architecture & Data Size & Avail. & Elo  \\ \midrule
Claude & unk. & unk. & API & 1215 \\
gpt-3.5-turbo & GPT3 & unk. & API & 1189 \\ \midrule
ChatGLM2-6B & GLM & 1.4TB* & Weights & 1104 \\
phoenix-chat-7B & BLOOMZ & 1.5GB & Full & 1065 \\
RWKV-4-World-7B & RWKV & 800GB* & Full & 1031 \\
Longjing-7B & T5 & unk. & Weights & 979 \\
\textbf{ChatFlow (ours)} & LLaMA & 50GB & Full & 868 \\
RWKV-4-Raven-7B & RWKV & 800GB* & Full & 852 \\
baichuan-7B & LLaMA & 1.2T* & Weights & 816 \\
LMFlow-Robin-7B & LLaMA & 500MB & Full & 658 \\ \bottomrule
\end{tabular}}
%\vspace{-0.5cm}
\end{table}

We conduct an ablation study to analyze the impact of our proposed dynamic data sampler. For the baseline, we train a naive Chinese-LLaMA2, utilizing the same data and hyperparameters as those used in ChatFlow, but without the the dynamic data sampler. We track the metrics throughout the training process. Figure~\ref{loss} illustrates the convergence of the training loss, indicating that ChatFlow demonstrates a lower loss at the beginning of the training and converges more swiftly. In Figure~\ref{exp1}, we evaluate the model's performance on the MMLU and C-Eval benchmarks. The results reveal that at each training stage, the model with the dynamic data sampler consistently achieves more stable and superior scores.

\section{Conclusion}

In this paper, we propose ChatFlow to address the challenge of training cross-language LLMs in a cost-effective manner. By leveraging a mix of Chinese, English, and parallel corpora, we align cross-language representations and facilitate knowledge transfer to the Chinese language model. In addition, we introduce a dynamic data sampler to train ChatFlow with a smooth transition from unsupervised pre-training to supervised fine-tuning. We train ChatFlow using our proposed protocol, it serves as a reproducible baseline and valuable reference for transferring to other languages. We have publicly made our code and weights available to ensure reproducibility and encourage further research.

\section{Acknowledgement}

This work was supported by the National Natural Science Foundation of China under Grant 82261138629, 62206180; Guangdong Basic and Applied Basic Research Foundation under Grant 2023A1515010688, 2022A1515011018; Shenzhen Municipal Science and Technology Innovation Council under Grant JCYJ20220531101412030 and XJTLU RDF under Grant RDF-23-01-053.

%Moving forward, 

\vfill\pagebreak

\bibliographystyle{IEEEbib}
\bibliography{strings,refs}

\end{document}